\documentclass[runningheads]{llncs}
\usepackage[T1]{fontenc}
\usepackage{graphicx}
\usepackage{booktabs}
\usepackage[misc]{ifsym}

\usepackage{mwe}
\usepackage{multirow} 
\usepackage{amssymb}
\usepackage{amsmath}
\usepackage{subcaption}
\usepackage{graphicx}
\usepackage{hyperref}
\usepackage{xcolor}
\usepackage{cleveref}

\usepackage{bm}

\def\T{{\mathcal{T}}}
\def\H{{\mathbf{H}}}
\def\h{{\mathbf{h}}}
\def\R{{\mathbb{R}}}
\def\L{\mathcal{L}}
\def\U{\mathbf{U}}
\def\V{\mathbf{V}}
\def\K{\mathbf{K}}
\def\W{\mathbf{W}}
\def\u{\mathbf{u}}

\def\w{\mathbf{w}}
\def\M{\mathbf{M}}
\def\A{\mathbf{A}}
\def\E{\mathbf{E}}

\def\X{\mathbf{X}}
\def\P{\mathbf{P}}
\def\p{\mathbf{p}}
\def\Z{\mathbf{Z}}

\begin{document}

\title{Dual-Attention Convolution Experts for Sparse Tensor Completion}


\titlerunning{Dual-Attention Convolution Experts for Sparse Tensor Completion}

\author{Yanlei Liu \and Zhenyu Liao\Letter}
\authorrunning{Y. Liu and Z. Liao}
\institute{
School of Electronic Information and Communications (EIC),\\
Huazhong University of Science and Technology (HUST), Wuhan, China\\
\email{M202473156@hust.edu.cn, zhenyu\_liao@hust.edu.cn}
}
\tocauthor{Yanlei Liu, Zhenyu Liao}
\toctitle{Dual-Attention Convolution Experts for Sparse Tensor Completion}

\maketitle              

\begin{abstract}
Tensor factorization (TF) has been widely adopted for high-dimensional sparse data completion tasks. 
Despite significant progress, neural TF methods often struggle to capture complex cross-mode interactions and remain vulnerable to (extreme) data sparsity.
To address these challenges, we propose a novel neural tensor factorization approach, termed \textbf{D}ual-Attention \textbf{C}onvolution Expert Networks with \textbf{G}roup-Level \textbf{C}ontrastive Learning (\textbf{DCGC}). 
For the first problem, DCGC generates diverse non-linear alignment patterns of latent factors via a multi-channel convolution network, and leverages the gated dual-attention mechanism to drive the model to focus on more important output channels (i.e., convolution experts) and the aligned features. 
Furthermore, DCGC introduces a group-level contrastive learning strategy that aggregates positive samples with identical feedback levels while separating negative samples across different levels. 
This strategy injects high-quality self-supervised signals to mitigate data sparsity.  
Extensive experiments conducted on five datasets demonstrate that our DCGC outperforms the state-of-the-art methods in sparse tensor completion for traffic and recommendation applications. 
Code to reproduce the experimental results in the paper is available at \url{https://github.com/ku1z/DCGC}.
 
\end{abstract}

\section{Introduction}

Tensors provide a natural and powerful representation for multidimensional data arising in many real-world applications. 
Examples include recommendation systems, spatiotemporal data analysis, and image processing. 
However, real-world tensors are typically extremely \emph{sparse} and \emph{incomplete}. 
For instance, in context-aware recommendation systems, a triplet $(\text{user}, \text{item}, \text{context})$ in a rating tensor records the rating given by a user to an item under a specific context. 
Since each user interacts with only a small subset of items, the resulting tensor is highly sparse \cite{li2025coatf}.  
Tensor factorization (TF), which aims to recover missing entries from partially observed tensors, has therefore become a fundamental tool for modeling such data and has been widely applied to tasks such as spatiotemporal forecasting \cite{wang2023spatialtemporal2}, recommendation \cite{wu2017cars1}, and image denoising \cite{imagedenoising2}.

A central challenge in TF is how to effectively model complex cross-mode interactions through latent factors \cite{liu2019costco}. 
Traditional tensor decomposition methods \cite{yuan2019tensorring} primarily rely on linear or multi-linear operations, which limits their ability to capture highly \emph{nonlinear} relationships in modern datasets. 
Motivated by the success of deep learning, neural tensor factorization approaches have attracted growing interest recently, due to their ability to learn expressive nonlinear representations. 
Existing nonlinear models can generally be divided into two main categories: MLP-based methods \cite{dai2024hoctc,fan2022MDMTF} and CNN-based architectures \cite{li2025coatf,NTC}. 
While CNN-based models often reduce the risk of overfitting through parameter sharing \cite{liu2019costco}, their convolutional structures may still struggle to capture fine-grained cross-mode interactions among latent factors.

Another fundamental difficulty arises from the severe \emph{sparsity} of observed tensor entries \cite{sun2021datasparsity}. 
This problem becomes even more pronounced when deep neural networks are employed, as complex models require sufficient model capacity as well as sufficiently strong training signals. 
Several approaches attempt to alleviate this issue by incorporating auxiliary information. 
For example, \cite{ioannidis2019coupled} incorporates auxiliary information by coupling similarity matrices, \cite{wang2023biastr} aggregates information from neighboring nodes within social networks, and \cite{li2025coatf} leverages generalized implicit feedback to comprehensively represent user preferences. 
Nonetheless, these approaches are often tailored to specific application scenarios and may introduce additional noise when auxiliary information is unreliable. 
Consequently, designing nonlinear tensor factorization models that remain robust under sparsity remains an open challenge.

To address these issues, we propose \textbf{D}ual Attention \textbf{C}onvolution Expert Networks with \textbf{G}roup-Level \textbf{C}ontrastive Learning, \textbf{DCGC}, a novel neural-network-based framework for \emph{efficient and sparse} tensor completion. 
DCGC explicitly models higher-order cross-mode interactions while improving robustness to data sparsity. 
We first construct second-order cross features by introducing pairwise Hadamard interactions between latent factors.  
Unlike existing CNN-based approaches, we formulate convolution channels as a fine-grained mixture-of-convolution-experts and employ a dual-attention mechanism to dynamically prioritize critical experts and aligned features. This design enables DCGC to maintain competitive performance even with a low rank and a reduced number of convolutional output channels.
In addition, we introduce a personalized gating strategy to improve expert load
balancing and encourage the model to focus on latent factors with limited interactions, thereby enhancing performance in data-sparse scenarios. 
To further alleviate data sparsity, we develop a group-level contrastive learning strategy that aggregates samples with identical feedback levels while separating those with different feedback levels in the representation space, thereby providing high-quality self-supervised signals.


\medskip

The primary contributions of this paper are summarized as follows.
\begin{enumerate}
    \item We propose a \textbf{mixture-of-convolution-experts network} with a dual-attention mechanism and personalized gating, which enables fine-grained modeling of nonlinear cross-mode interactions among tensor latent factors.

    \item We develop an efficient \textbf{in-batch group-level contrastive learning strategy} that exploits feedback-level consistency to provide high-quality self-supervised signals and alleviate the impact of data sparsity.

    \item We conduct extensive experiments on five real-world datasets ranging from traffic prediction to recommendation tasks. 
    The empirical results demonstrate that the proposed method consistently outperforms state-of-the-art TF approaches such as CoATF \cite{li2025coatf} and HOCTC \cite{dai2024hoctc}. 
\end{enumerate}

\section{Related Work}

\paragraph{Tensor Factorization.}
Tensor factorization aims to reconstruct a tensor from its observed entries using learned latent factors. 
Multi-linear tensor decomposition has a long-standing history, evolving from classical approaches such as CP factorization \cite{harshman1970cp} and Tucker factorization \cite{tucker1966tucker}, to more recent advancements like Tensor Ring (TR) \cite{yuan2019tensorring}, which connects adjacent modes in a circular chain. Recently, neural network-based tensor factorization methods have attracted increasing attention due to their robust non-linear modeling capabilities \cite{wu2019neuraltensor}. For instance, \cite{fan2022MDMTF} utilizes MLPs to model non-linear latent factors within Tucker factorization, while \cite{liu2019costco} leverages the expressive power of CNNs to align different factors non-linearly. 
Furthermore, to capture complex structural dependencies, \cite{wang2023dynamictensor} employs a neural diffusion-reaction process over a multi-partite graph.

\paragraph{Contrastive Learning.}
Contrastive learning is a powerful self-supervised paradigm for learning feature embeddings that are discriminative and generalizable. In recommendation system, \cite{zhang2025unveiling} introduces a contrastive objective that enables efficient neighborhood aggregation. In natural language processing, \cite{gao2021nlp_ssl} performs contrastive learning by using dropout as minimal noise to align a sentence with itself, whereas in computer vision, \cite{chen2023cvssl} derives pseudo-semantic labels from unlabeled human images to define positive and negative pairs for self-supervised training. In addition, \cite{assran2023cvssl1} learns semantic image representations by predicting embeddings of large masked target blocks from a spatially informative context block, thereby eliminating the need for handcrafted augmentations.



\paragraph{Data Sparsity in Recommendation System.} 
Existing studies targeting data sparsity in recommendation systems also motivate the design of a more general tensor factorization framework. For instance, \cite{yao2021google_ssl} propose a multi-task self-supervised learning approach for large-scale item recommendation.
To mitigate the inherent user cold-start challenge, \cite{dai2021poso} introduce a novel gating strategy. 
Furthermore, \cite{lu2020metahin} integrate heterogeneous information networks with a meta-learning paradigm to capture complex network semantics. 
Similarly, \cite{cao2023coldgpt} address data sparsity by pre-training an extendable item-attribute graph via multi-task learning across diverse data sources. 
Recently, \cite{lei2025fsgnn} tackle the cold-start recommendation problem by adaptively combining LLM-enhanced feature completion and three-channel structure augmentation to generate robust node embeddings.






\section{Method}

This section presents the proposed DCGC model in detail. \Cref{sub:model_overview} formulates the tensor factorization (TF) task and outlines the overall framework. Subsequently, \Cref{sub:model_stucture} describes the primary architecture of the model, which comprises an embedding layer that explicitly incorporates second-order interactions among latent factors, alongside a nonlinear mapping module. This mapping module consists of a mixture of convolutional experts to effectively align distinct latent factors, a dual-attention mechanism that highlights critical experts and aligned features, and a personalized gating module designed to prevent attention polarization and mitigate data sparsity. Furthermore, \Cref{sub:multi_task_loss_training} introduces the group-level contrastive learning strategy, detailing the construction of positive and negative sample pairs and the overall loss function required for robust model optimization. Finally, \Cref{sub:complexity_analysis} analyzes the space and computational complexities of our proposed model.

\subsection{Model Overview}\label{sub:model_overview}

Without loss of generality, we illustrate the proposed DCGC model using a third-order tensor.
Let $\T \in \R^{I \times J \times K}$ denote a partially observed tensor, and let $\mathcal{I} \subseteq [I]\times[J]\times[K]$ be the set of observed entries.
The goal of TF is to accurately recover the missing entries of $\T$ from the observed data.

Following the CP decomposition paradigm, we represent each mode using a rank-$R$ latent factor matrix that learns a prediction function
$f: [I]\times[J]\times[K] \rightarrow \R$, parameterized by factor matrices
$\U \in \R^{I\times R}$, $\P \in \R^{J\times R}$, $\W \in \R^{K\times R}$,
together with additional model parameters $\theta$.
Given an index triplet $(i,j,k)$, the predicted tensor entry is defined as
\begin{equation}\label{eq:gtf}
\hat{\T}_{ijk} = f(i,j,k; \U,\P,\W,\theta),
\end{equation}
where $f(\cdot)$ models the interaction among the corresponding latent factors.

In classic TF methods, the parameters are learned by minimizing the reconstruction error over the observed entries:
\begin{equation}
\min_{\U,\P,\W,\theta}
\sum_{(i,j,k)\in \mathcal{I}}
\left(\hat{\T}_{ijk}-\T_{ijk}\right)^2,
\end{equation}
where $\T_{ijk}$ denotes the $(i,j,k)$-th entry of $\T$.

\begin{figure}[t]
\includegraphics[width=\textwidth]{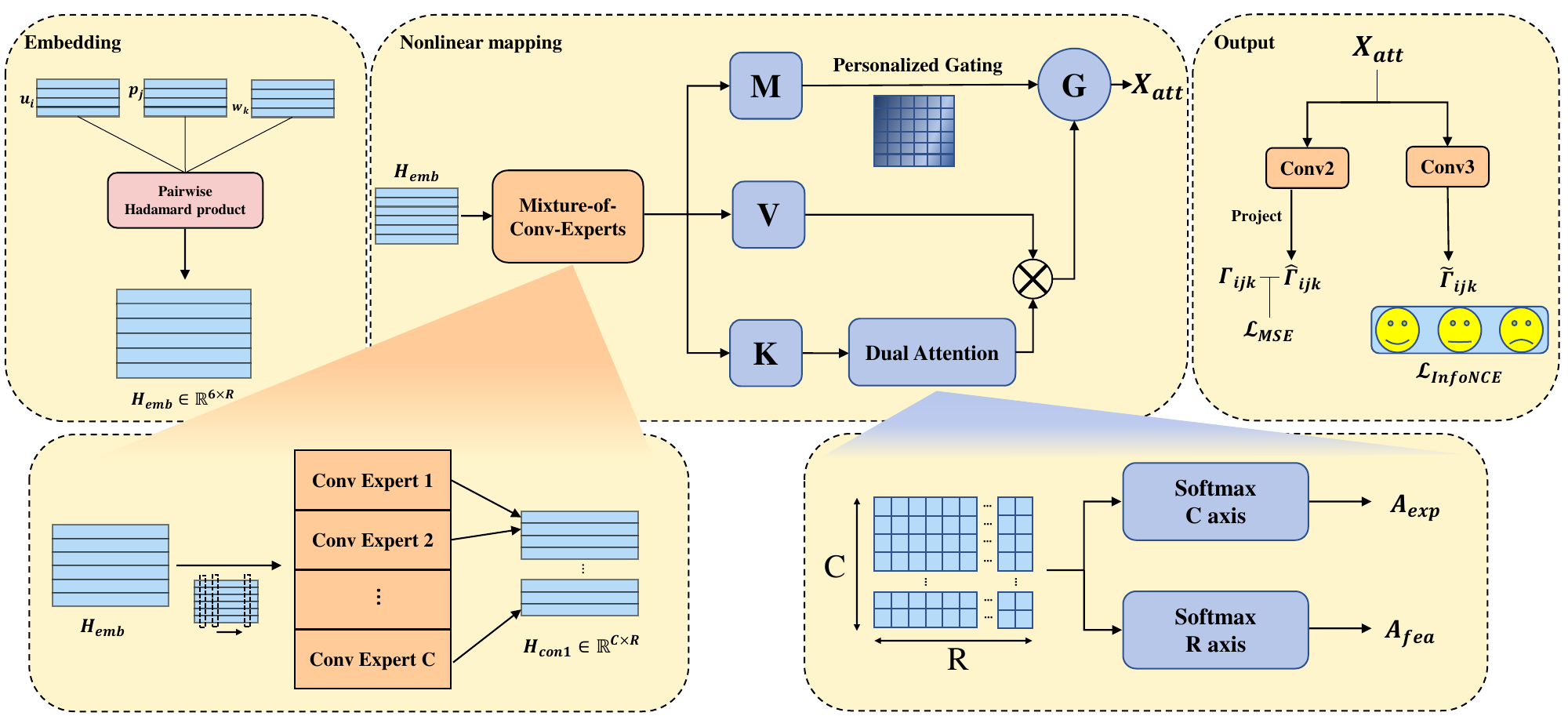}
\caption{The overall architecture of the DCGC framework. DCGC first constructs second-order interactions among latent factors through the embedding module, then models nonlinear cross-mode dependencies using a mixture of convolutional experts with dual-attention and personalized gating, and finally jointly optimizes tensor reconstruction and group-level contrastive learning objectives.} \label{fig:model_DCGC}
\end{figure}

In this paper, we propose DCGC, a novel neural TF framework to \eqref{eq:gtf}.
DCGC consists of three main components: an embedding layer, a mixture-of-convolution experts network, and a dual-branch output layer; see \Cref{fig:model_DCGC} for an overview.

\subsection{Model Structure}\label{sub:model_stucture}
This section describes the main architectural components of DCGC.

\paragraph{Embedding Module.} A key component in the integration of neural networks with tensor factorization models is the embedding module, which maps discrete indices to low-rank latent factors.
Given an observed triplet $(i,j,k)$ and its corresponding one-hot row vectors $\mathbf{a}_i \in \mathbb{R}^{1 \times I}$, $\mathbf{b}_j \in \mathbb{R}^{1 \times J}$, and $\mathbf{c}_k \in \mathbb{R}^{1 \times K}$, we obtain the dense, continuous embedding row vectors via linear projections: 
\begin{equation}
\mathbf{u}_i = \mathbf{a}_i \mathbf{U}, \quad \mathbf{p}_j = \mathbf{b}_j \mathbf{P}, \quad \mathbf{w}_k = \mathbf{c}_k \mathbf{W},
\end{equation}
where $\mathbf{U} \in \mathbb{R}^{I \times R}$, $\mathbf{P} \in \mathbb{R}^{J \times R}$, and $\mathbf{W} \in \mathbb{R}^{K \times R}$ are the learnable embedding matrices, and $R$ denotes the dimension of the latent factors.

To efficiently capture cross-dimensional relationships, we explicitly model the pairwise interactions between the latent factors using the element-wise Hadamard product $\odot$. 
These pairwise interaction vectors are then vertically stacked with the original embedding row vectors to construct a unified 2D representation:
\begin{equation}
{\H_{\rm emb}} = {\rm concat} [\u_i, \p_j, \w_k, \u_i \odot \p_j, \u_i \odot \w_k, \p_j \odot \w_k]\in\R^{6\times R}, 
\end{equation}
where ${\rm concat}[\cdot]$ denotes the vertical stacking of the row vectors. 
While this is a standard practice in classic multi-linear tensor decomposition \cite{rendle2010pairwise}, we argue that it remains equally effective within non-linear frameworks, particularly in our CNN-based architecture.

\paragraph{Mixture-of-Conv-Experts.} 
To mitigate negative transfer across tasks, inspired by the MMoE framework~\cite{ma2018mmoe}, we construct a mixture of convolutional experts, where each expert performs weighted nonlinear alignments of distinct latent factors. This multi-expert design enables individual experts to specialize in extracting heterogeneous underlying patterns, while mitigating overfitting through the sharing of convolutional parameters\cite{liu2019costco}. Shared across the primary and auxiliary tasks, these experts allow the model to absorb high-quality self-supervised signals while avoiding rigid representation entanglement.
\begin{equation}
    \left\{
    \begin{aligned}
    \h_{c} &= {\rm ReLU} \left(\text{Conv}(\H_{\rm emb}) \right),  \\
    \H_{\rm{con1}} &= {\rm{concat}}[\h_{1},\h_{2},\dots,\h_{C}]
    \end{aligned}
    \right.
\end{equation}
where $\h_{c} \in \mathbb{R}^{1\times R}$ denotes the output feature map of the $c$-th Conv expert, and $\H_{\rm{con1}}\in\mathbb{R}^{C\times R}$ represents the concatenated output from all $C$ experts stacked along the channel dimension. The indices are defined as $1 \leq c \leq C$ for the Conv experts, and $r$ denotes the index of the latent factor dimension. The convolution kernel size is set to $6\times 1$ to enable efficient alignment.

\paragraph{Dual-Attention.} Within the mixture-of-convolution-experts framework, we emphasize the necessity of modeling the importance of both individual convolution experts and aligned element-wise features. Specifically, we generate the gate, key, and value embeddings from $\H_{\text{con1}}$:
\begin{equation}
    \begin{aligned}
        \M,\K,\V = {\rm{Split}}(\text{SiLU}(f_1(\H_{\rm{con1}}))),
    \end{aligned}
\end{equation}
where $ \M, \K, \V \in \R^{C\times R}$ and $f_1$ denotes a learnable MLP. 
The activation function SiLU is employed to facilitate non-linear mapping. 

We then separately compute the attention scores for the convolution experts and the feature elements. Specifically, the expert attention $\mathbf{A}_{\text{exp}}$ is obtained by applying the softmax operation across the expert dimension (i.e., along the columns of $\mathbf{K}$), while the feature attention $\mathbf{A}_{\text{fea}}$ is computed across the feature dimension (i.e., along the rows of $\mathbf{K}$). The operations are formulated as follows:

\begin{equation}
    \begin{aligned}
    A^{\text{exp}}_{c,r} &= \frac{\exp(K_{c,r})}{\sum_{i=1}^{C}\exp(K_{i,r})}, \quad \text{for } c=1, 2, \dotsm, C \\
    A^{\text{fea}}_{c,r} &= \frac{\exp(K_{c,r})}{\sum_{j=1}^{R}\exp(K_{c,j})}, \quad \text{for } r=1, 2, \dotsm, R
    \end{aligned}
\end{equation}
where $K_{c,r}$ denotes the scalar element at the $c$-th row and $r$-th column of the key embedding. $\mathbf{A}_{\text{exp}} \in \mathbb{R}^{C \times R}$ and $\mathbf{A}_{\text{fea}} \in \mathbb{R}^{C \times R}$ denote the importance matrices corresponding to the convolution experts and feature elements, respectively, with their individual elements denoted as $A^{\text{exp}}_{c,r}$ and $A^{\text{fea}}_{c,r}$.

 As demonstrated in Section \ref{sec:para_ana}, by incorporating the dual-attention mechanism, our model achieves competitive performance even with a low rank and a reduced number of convolution experts. Subsequently, the attention-weighted outputs of the convolution experts $\E\in\R^{C\times R}$ are formulated as:
\begin{equation}
\label{eq:dual_attention_response}
    \E = \left(\A_{\rm{exp}} + \A_{\rm{fea}}\right) \odot \V .
\end{equation}
This additive fusion allows the value embedding $\V$ to be
modulated by both expert-level and feature-level importance.

\paragraph{Personalized Gating.} To optimally route the aforementioned convolution experts, we introduce a personalized gating mechanism addressing two major limitations of standard routing: inherent attention polarization and the neglect of sparsely observed latent factors:
\begin{equation}
\label{eq:personalized_gating}
\X_{\text{att}} = \text{SiLU}(f_2(\M \odot \E)),
\end{equation}
where $\X_{\text{att}} \in \R^{C\times R}$ is the output, $f_2$ is an MLP, and $\M$ denotes context-aware routing weights. By dynamically generating input-dependent $\M$ for each tensor entry, this tailored aggregation mitigates attention score polarization, where a few dominant convolution experts or features might otherwise overshadow others. It explicitly reallocates modeling capacity toward sparse interactions, demonstrating superior robustness over conventional dropout (Section \ref{sec:ablation_study}).

To further enhance model performance and ensure training stability, we apply a residual connection with layer normalization (LN) as follows:
\begin{equation}
    \begin{aligned}
    \H_{\rm out} &= \rm{}LN\left(\X_{\rm{}att}+\H_{\rm{}con1}\right)\in \R^{C\times R}.
    \end{aligned}
\end{equation}

\paragraph{Output.} 
Building upon the shared bottom architecture, our model diverges into two functional branches: one dedicated to the primary task of recovering missing tensor values, and the other serving as an auxiliary task to mine structural information within the tensor. The output of the main task is obtained by:
\begin{equation}
\begin{aligned}
\hat{\T}_{ijk} = f_3(\text{ReLU}(\text{Conv}(\H_{\text{out}}))),
\end{aligned}
\end{equation}
where $\text{Conv}$ denotes a convolution layer with a kernel size of $1 \times R$, and $f_3$ refers to the output mapping layer. 
The auxiliary prediction $\widetilde{\T}_{ijk}$ is obtained identically.

\subsection{Group-Level Contrastive Learning}\label{sub:multi_task_loss_training}
A key challenge in tensor decomposition lies in \emph{data sparsity}, an issue that becomes particularly exacerbated within neural tensor decomposition models. By introducing Group-Level Contrastive Learning (GLCL), we aim to leverage latent factors possessing relatively abundant interaction information to facilitate the representation learning of those with sparse interactions. In this section, we provide the detailed design of GLCL and our multi-task learning framework.

\paragraph{Group-Level Contrastive Learning.}

\begin{figure}[t]
\includegraphics[width=\textwidth]{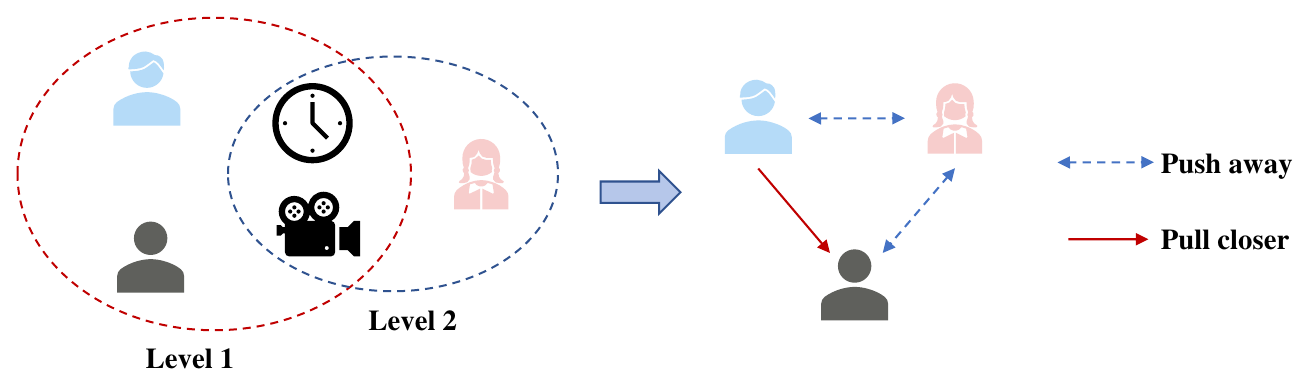}
 \caption{Illustration of our proposed Group-Level Contrastive Learning strategy in recommendation scenario. GLCL enforces intra-group aggregation and inter-group separation via contrastive learning to alleviate data sparsity.} \label{fig:GLCL}
\end{figure}
First, we categorize the triplets $(i,j,k)$ into distinct feedback levels based on original tensor values. Let $\mathcal{\omega}=[\mathcal{\omega}_1, \mathcal{\omega}_2, \dots, \mathcal{\omega}_l]$ be an ordered set of feedback levels, where $\mathcal{\omega}_{p+1}>\mathcal{\omega}_{p}$ indicates that feedback in $\mathcal{\omega}_{p+1}$ represents a stronger signal of interest than feedback in $\mathcal{\omega}_{p}$. Taking the MovieLens movie rating dataset as an example, where the rating scale ranges from 1 to 5, we partition the ratings into three distinct feedback levels:
\begin{equation}
    \begin{aligned}
        \text{Like}(5\sim4):\mathcal{\omega}_3\quad \text{Neutral}(3):\mathcal{\omega}_2\quad
        \text{Dislike}(2\sim1):\mathcal{\omega}_1
    \end{aligned}
\end{equation}

In this study, to enhance the generalizability and applicability of the model, we partitioned each spatiotemporal tensor into three feedback intervals of equal width according to its original value range. Specifically, taking the GZspeed tensor with values ranging from $0$ to $120$ as an example, we uniformly divided its value space into three intervals:
\begin{equation}
    \begin{aligned}
        \text{High}(120\sim80):\mathcal{\omega}_3\quad \text{Medium}(80\sim40):\mathcal{\omega}_2\quad
        \text{Low}(40\sim0):\mathcal{\omega}_1
    \end{aligned}
\end{equation}

As illustrated in Figure \ref{fig:GLCL}, active users with rich behavioral histories can transfer informative auxiliary signals to cold-start users in recommendation systems through intra-group aggregation and inter-group isolation. The same principle applies to spatiotemporal tensors: latent factors corresponding to low-activity sensors are inadequately updated during training, whereas high-activity sensors located in the same climate zone or assigned to the same road category share comparable environmental attributes and functional roles. By integrating GLCL, these sparse sensors are clustered with high-activity counterparts that exhibit similar environmental attributes or functional roles, e.g. sensors situated in the same climate zone or assigned to the same road category. Consequently, the abundant information from active regions is utilized to stabilize the representation learning of the sparse sensors.

For a given observed triplet $(i,j,k)$, we define its negative samples $(i',j',k')\in \mathcal{N}$ as follows: 
\begin{enumerate}
    \item $(i',j',k')$ shares at least one common index with $(i,j,k)$, i.e., $i=i'$ or $j=j'$ or $k=k'$; and 
    \item the feedback level of $(i',j',k')$ is lower than that of $(i,j,k)$, i.e., $\mathcal{\omega}^{ijk}>\mathcal{\omega}^{i'j'k'}$. 
\end{enumerate}

Then, we introduce InfoNCE loss to incorporate abundant self-supervised signals through neighborhood information aggregation on the interaction graph \cite{zhang2025unveiling}, which is an effective approach to incorporate graph-structured information into tensor-based methods. Our GLCL loss is defined as:
\begin{equation}
    \begin{aligned}
        \L_{\rm CL} = -\log\left(\frac{\exp(\widetilde{\T}_{ijk}/\tau)}{\exp(\widetilde{\T}_{ijk}/\tau)+\sum_{(i'j'k')\in \mathcal{N}}\exp(\widetilde{\T}_{i'j'k'}/\tau)}\right),
    \end{aligned}
\end{equation}
where $\tau$ is a hyperparameter that governs the model's discriminability towards negative samples. 

In summary, by integrating feedback level grouping with the InfoNCE loss, we have developed a group-aware contrastive learning strategy. As demonstrated in Sec \ref{sec:ablation_study}, our GLCL strategy effectively mitigates data sparsity and prevents the introduction of noisy signals that could potentially interfere with the optimization of the primary task, outperforming traditional contrastive methods.

\paragraph{Multi-task Joint Training.}

To reconstruct the original input tensor as accurately as possible, we still employ Mean Squared Error (MSE) as the primary loss to guide the optimization direction, which is defined as:
\begin{equation}
    \begin{aligned}
        \L_{\rm main} = \frac{1}{2}
\sum_{(i,j,k)\in \mathcal{I}}
\left(\hat{\T}_{ijk}-\T_{ijk}\right)^2.
    \end{aligned}
\end{equation}
where $\T_{ijk}$  denotes the ground-truth values of the original tensor, and $\hat{\T}_{ijk}$ denotes the tensor reconstructed by the model.

Through multi-task joint learning, we introduce the contrastive learning loss to mitigate data sparsity. 
Then, the loss of the proposed framework is given by:
\begin{equation}
    \begin{aligned}
        \mathrm{\L = \L_{main} + \alpha\L_{CL} + \beta\L_r(\U,\P,\W,\theta)},
    \end{aligned}
\end{equation} 
where $\alpha$ denotes the weight of the auxiliary task and $\L_r$ is the $L2$ regularization loss. 
We recommend setting $\alpha$ to be relatively small, e.g., 0.2 $\sim$ 0.5, to ensure that the main task $\rm\L_{\rm main}$ dominates the initial training phase. Once the main task becomes relatively stable, the auxiliary task intervenes and introduces high-quality self-supervised signals to alleviate data sparsity.

\subsection{Complexity Analysis}\label{sub:complexity_analysis}

\paragraph{Space Complexity.} 

Let $R$ and $C$ denote tensor rank and the number of convolution experts. 
In DCGC, the learnable parameters mainly come from the embedding layer. 
In addition, the number of parameters in the non-linear mapping\footnote{We ignore the bias terms of the convolutional and fully-connected layers.} is $6C+4R^2$, and the number of parameters in the output is $2\left(C^2R + C\right)$. 
The space complexity of the embedding layer is $\left(I+J+K\right)R$. Given that $\left(I+J+K\right) \gg C,R$, the space complexity of DCGC is $\mathcal{O}\left(\left(I+J+K\right)R\right)$, which is similar to other methods based on the CP decomposition paradigm.

\paragraph{Time Complexity.} 
Let $B$ represent the size of the batch. 
DCGC adopts an efficient softmax operation for computing attention scores, with a computational complexity of $\mathcal{O}\left(B\left(R\times C\right)\right)$. 
We adopt a sampling method similar to \cite{dai2024hoctc} for group-level contrastive learning, where the time complexity is proportional to $B$. 
Therefore, the computational cost of training DCGC is mainly concentrated on the $\M,\K,\V$ embedding projection, Conv2 and Conv3. 
The time complexity of embedding projection is $\mathcal{O}\left(B\left(C\times R^{2}\right)\right)$. 
The time complexity of Conv2 and Conv3 is $\mathcal{O}\left(B\left(C^{2}\times R\right)\right)$. Hence, the time complexity of DCGC is $\mathcal{O}\left(B\left(CR^{2}+C^{2} R\right)\right)$. 
Notably, as shown in Sec \ref{sec:para_ana}, DCGC typically requires a smaller number of convolution channels, so its computational complexity is comparable to that of other CNN-based methods.

\section{Experiments}
In this section, we present a systematic experimental evaluation to validate our model's effectiveness. This includes comparative performance analysis in two distinct scenarios, ablation studies to isolate the contribution of key modules, and hyperparameter sensitivity analyses to demonstrate the model’s robustness.

\subsection{Experiment Setup}
\begin{table}[t]
    \caption{Statistics of the Datasets}
    \label{tab:datasets}
    \centering
    \begin{tabular}{c c c c c c}
    \toprule
    Dataset        & PEMS   & GZspeed    & Mov100k  & Mov1m     & Beauty\\
    \midrule
    Dimension1     & 100        & 214        & 943      & 6040      & 324038 \\
    Dimension2     & 181        & 61         & 1682     & 3706      & 32586  \\
    Dimension3     & 288        & 144        & 7        & 30        & 30     \\
    Observed       & 521,160    & 187,977    & 100,000  & 1,000,000 & 371,345 \\
    Sampling ratio & 10.0$\%$   & 10.0$\%$   & 6.3$\%$  & 4.47$\%$  & 0.0035$\%$ \\
    \bottomrule
    \end{tabular}
\end{table}

\paragraph{Datasets.}


To verify the effectiveness of our method, we select spatiotemporal and recommendation tensors, both popular benchmarks in previous work \cite{wang2023biastr}. 
We consider two traffic datasets (GZspeed, PEMS) and three recommendation datasets (Movielens-100k, Movielens-1M, AmazonBeauty). 
Following \cite{dai2024hoctc}, we randomly sample $10\%$ of entries for traffic datasets to evaluate completion algorithms. These are formatted as (sensor id, day id, interval id) triplets. 
For recommendation tensors, we use ``timestamp'' as context. 
Following \cite{li2025coatf}, we map timestamps in Movielens-100k to seven days of the week, and to 31 days of the month for Movielens-1M and AmazonBeauty. These follow a (user id, item id, context id) triplet format. \Cref{tab:datasets} provides further dataset details.

\paragraph{Baselines.}
To evaluate the proposed method, we compare it against P-Tucker \cite{oh2018ptucker}, a classic linear tensor decomposition method, and five recent neural nonlinear methods. 
These include CNN-based CoSTCo \cite{liu2019costco} and CoATF \cite{li2025coatf}, alongside MLP-based NTM \cite{chen2020ntm}, MDMTF \cite{fan2022MDMTF}, and HOCTC \cite{dai2024hoctc}. 
Notably, in context-aware recommendation, time represents just one context type supported by our framework. 
Following CoATF \cite{li2025coatf}, we do \emph{not} treat timestamps as dynamic features with explicit sequential order. 
Instead, we encode them as static features (e.g., day of the week or month) to ensure the model remains applicable to wider contextual scenarios. Consequently, methods designed specifically for sequential recommendation are excluded from the baseline comparison.
\paragraph{Implementation Details.}  
All experiments are conducted on a single NVIDIA RTX 3090 GPU. 
The parameters for the baselines are initialized as in the original papers and carefully tuned to achieve optimal performance. 
For our method, we set the tensor rank to 30 for DCGC and all baselines. 
We employed a grid search approach to find appropriate parameters, with the batch size selected from $\left\{128,256,512,1024\right\}$, learning rate within $\left\{0.0001,0.001,0.003,0.01\right\}$ and the weight of the auxiliary task $\alpha$ within $\left\{0.2,0.3,0.4,0.5\right\}$. 
To evaluate the model performance under different data sparsity levels, we compare the results of different baselines with training set ratios of 80$\%$ and 90$\%$, respectively. 
Following \cite{li2025coatf}, we adopt two standard metrics for evaluation: Mean Absolute Error (MAE) and Root Mean Square Error (RMSE).

\subsection{Evaluation on spatiotemporal and recommendation tensors}

To comprehensively evaluate the effectiveness of our proposed model, we compare it against various tensor factorization methods across five datasets spanning two distinct scenarios: traffic and recommendation systems.
Furthermore, we incorporate several GNN-based methods specifically designed to address the data sparsity challenge within recommendation tasks.
Baseline results for tensor factorization methods are partially adopted from \cite{li2025coatf} and \cite{dai2024hoctc}, whereas those for GNN-based methods are directly sourced from \cite{lei2025fsgnn}.
\begin{table*}[!t]
  \caption{\small{Tensor completion results of spatiotemporal tensors. The best results are shown in \textbf{bold}, and the second-best results are \underline{underlined}.}}
  \label{tab:res_spatiotemporal}

  \centering
\scalebox{0.9}{
    \begin{tabular}{lccccccccc}
    \toprule                 
    Dataset     & Training ratio & Metrics & P-Tucker  & NTM  & CoSTCo    & MDMTF &HOCTC & CoATF   & \textbf{DCGC} \\
    \midrule
    \multirow{4}{*}{GZspeed}   & \multirow{2}{*}{80$\%$ }   & RMSE    & 4.8718 & 4.4097 & 4.6042 & 4.5414 & 4.4491 & \underline{4.3798} &   \textbf{4.2252}  \\
    
    ~ & ~ & MAE & 3.2827 & 2.8758 & 3.0118 & 3.0411 & 2.8937& \underline{2.8645} &  \textbf{2.7936} \\

    ~ & \multirow{2}{*}{90$\%$ }& RMSE & 4.8474 & 4.3690&  4.5328 & 4.5309& 5.3753 & \underline{4.3674} & \textbf{4.1348} \\

    ~ & ~ & MAE& 3.2718 & 2.8288 & 2.9860  & 3.0023 &3.6258 & \underline{2.8330} & \textbf{2.7492}\\
    \midrule
    \multirow{4}{*}{PEMS}   & \multirow{2}{*}{80$\%$ }   & RMSE    & 5.2395 & 4.1611 & 4.5304 & 4.8697 & 4.9861& \underline{4.0760} &   \textbf{3.9594}   \\
    
    ~ & ~ & MAE & 3.1032 & 2.2607 & 2.4140 & 2.6855 & 2.8173& \underline{2.1848} &  \textbf{2.0996} \\

    ~ & \multirow{2}{*}{90$\%$ }& RMSE &5.1164 & \underline{3.9806} & 4.4344 & 4.6690&4.8513 & 4.0656 &\textbf{3.8598} \\

    ~ & ~ & MAE& 2.8860 &\underline{2.1242}& 2.3967 & 2.5779 &2.7210& 2.1651 &\textbf{2.0928} \\
    \bottomrule
  \end{tabular}
  }

\end{table*}
\begin{table*}[!t]

  \caption{\small{Tensor completion results of recommendation tensors. The best results are shown in \textbf{bold}, and the second-best results are \underline{underlined}.}}
    \label{tab:res_rec}

  \centering
\scalebox{0.9}{
    \begin{tabular}{lccccccccc}
    \toprule                 
    Dataset     & Training ratio & Metrics & P-Tucker  & NTM  & CoSTCo    & MDMTF &HOCTC & CoATF   & \textbf{DCGC} \\
    \midrule
    \multirow{4}{*}{Mov100k}   & \multirow{2}{*}{80$\%$ }                & RMSE    & 1.1928 &0.9417 & 0.9461 & \underline{0.9365} & 0.9655& 0.9429  &   \textbf{0.9062}  \\
    
    ~ & ~ & MAE & 0.8225 & 0.7384 & 0.7409  & \underline{0.7374} &0.7653& 0.7431 & \textbf{0.7096}  \\

    ~ & \multirow{2}{*}{90$\%$ }& RMSE & 1.1664&0.9438 & 0.9386  & \underline{0.9331} &0.9636 & 0.9387 &\textbf{0.8999} \\

    ~ & ~ & MAE&0.8087 & 0.7386 & \underline{0.7343}  &  0.7356 &0.7619& 0.7438 &\textbf{0.7054}\\
    \midrule
    \multirow{4}{*}{Mov1M}   & \multirow{2}{*}{80$\%$ }                & RMSE      & 1.0754 & 0.9283  & 0.9487 &  0.9216 & \underline{0.8936}& 0.9342  & \textbf{0.8769}\\
    
    ~ & ~ & MAE & 0.7513 & 0.7236 & 0.7456  & 0.7320&\underline{0.7013} & 0.7392 & \textbf{0.6885} \\

    ~ & \multirow{2}{*}{90$\%$ }& RMSE & 0.9818 & 0.9208 &0.9395  & 0.9194 &\underline{0.8865}& 0.9266 & \textbf{0.8737}\\

    ~ & ~ & MAE & 0.7247 & 0.7169&0.7363  & 0.7287 & \underline{0.6960}&0.7327 & \textbf{0.6868}\\
    \midrule
    \multirow{4}{*}{Beauty}   & \multirow{2}{*}{80$\%$ }                & RMSE      & 2.9625 & 1.3396  & 1.3252  & \underline{1.2964} & 1.4296 &1.2998  & \textbf{1.2698}  \\
    
    ~ & ~ & MAE & 2.3804 &1.0818 & 1.0257  & \underline{0.9724} & 1.1349 & 0.9851 & \textbf{0.9001} \\

    ~ & \multirow{2}{*}{90$\%$ }& RMSE&2.7330 & 1.3212 & \underline{1.2880}  &1.2935&1.5259& 1.2970  &  \textbf{1.2607}\\

    ~ & ~ & MAE & 2.1581 & 0.9842 & 1.0234  & \underline{0.9510} & 1.1095&0.9830 & \textbf{0.8845}\\   
    \bottomrule
  \end{tabular}
  }
\end{table*}

\paragraph{Spatiotemporal Tensors.} Table \ref{tab:res_spatiotemporal} shows the tensor completion results on two spatiotemporal datasets. Notably, traffic tensors are inherently complete with simple internal data trends, so CNN-based methods with relatively few parameters, such as CoATF, often achieve better performance. In terms of RMSE and MAE, our proposed method outperforms the most competitive baseline by $3.34\%\sim5.06\%$ on the two datasets. Compared with the traditional linear method P-Tucker, the performance gains are even more substantial, reaching $13.27\%\sim32.34\%$.

\paragraph{Recommendation Tensors.}
Table \ref{tab:res_rec} shows the tensor completion results on three recommendation datasets. Recommendation tensors typically suffer from inherently sparse user-item-context interactions and complex dependency structures, where MLP-based methods, such as HOCTC and MDMTF, generally yield superior results. 
Compared to the second-best result, our model has the highest improvement rate of 3.56$\%$ and the lowest of 1.44$\%$ for the RMSE metrics and the highest of 7.43$\%$ and the lowest of 1.32$\%$ for the MAE metrics. 
\begin{figure}[t]
\includegraphics[width=\textwidth]{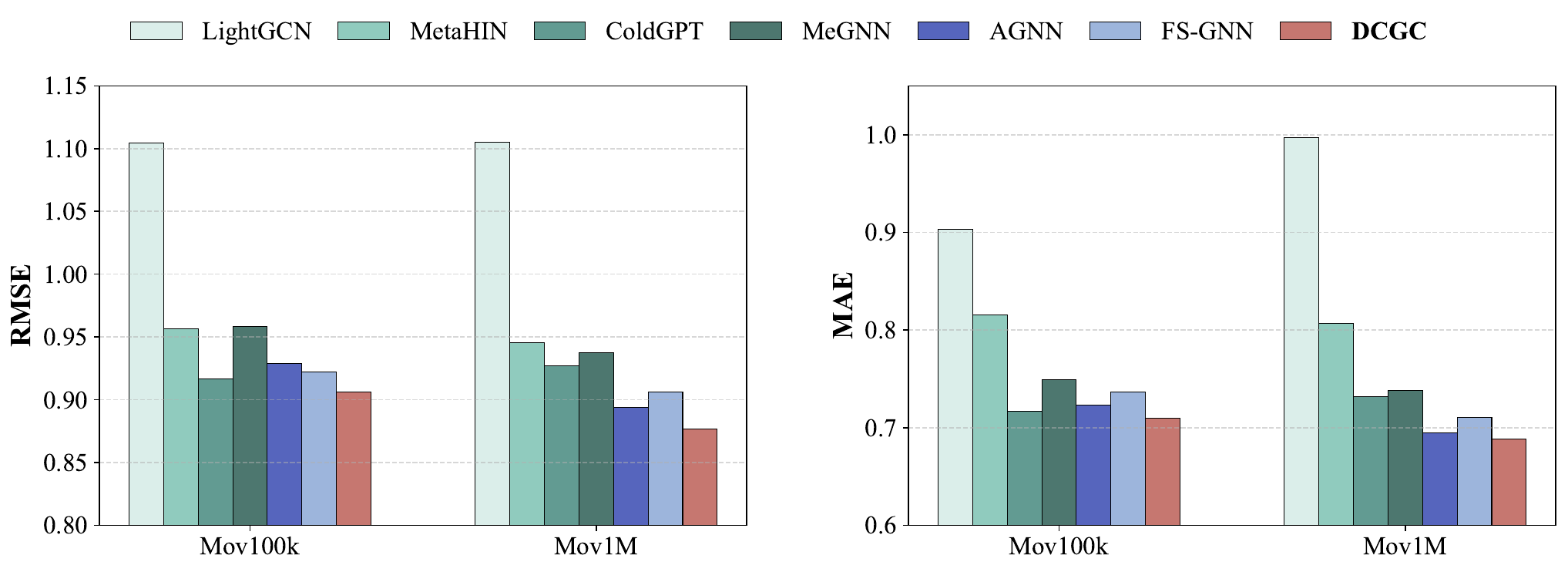}
\caption{Comparison with GNN-based methods designed for cold-start scenarios on recommendation datasets, where DCGC still achieves strong performance.} 
\label{fig:com_gnn}
\end{figure}

To further evaluate our model's performance on recommendation datasets, we also select several GNN-based methods LightGCN \cite{he2020lightgcn}, MetaHIN \cite{lu2020metahin}, ColdGPT \cite{cao2023coldgpt}, MeGNN \cite{liu2023megnn}, AGNN \cite{qian2020agnn} and FS-GNN \cite{lei2025fsgnn}, that also focus on solving data sparsity in recommendation systems for comparison. As shown in Figure \ref{fig:com_gnn}, compared with the most competitive SOTA GNN-based baseline, our proposed DCGC still achieves competitive and consistently superior performance.

\subsection{Ablation Study}
\begin{table}[!t]
    \centering
    \caption{Ablation study of the proposed components. We evaluate the contribution of each module by removing or replacing individual components.}
    \label{tab:ablation}
    \begin{tabular}{ccccccccccc}
    \toprule
    \multirow{2}{*}{Setting}  &  \multicolumn{2}{c}{Mov100k}  & \multicolumn{2}{c}{Mov1M} & \multicolumn{2}{c}{Beauty}   & \multicolumn{2}{c}{GZspeed} & \multicolumn{2}{c}{PEMS}\\ 
    ~ &   RMSE   &   MAE   &  RMSE  & MAE & RMSE & MAE & RMSE & MAE& RMSE & MAE \\
    \midrule
    \textbf{DCGC} & 0.9062 & 0.7096 & 0.8769 & 0.6885 &1.2698 & 0.9001 & 4.2252 & 2.7936 & 3.9594 & 2.0996\\
    \midrule
    FirstOrder & 0.9134 & 0.7163 & 0.8781 & 0.6905 & 1.2717 & 0.9031 & 4.2950& 2.8308 & 3.9904 & 2.1453\\
    \midrule
    ChannelAtt   & 0.9208 & 0.7215 & 0.8776 & 0.6899 & 1.2726 & 0.9140 & 4.2559 & 2.8039 & 3.9650 & 2.1332 \\
    FeatureAtt & 0.9075 & 0.7104 & 0.8966 & 0.7058 & 1.2709 & 0.9147 &
    4.2481 & 2.8144 & 3.9644 & 2.1239\\
    NoAtt &  0.9237  &  0.7212 & 0.8987 & 0.7072 & 1.2738 & 0.9296 &  4.3029 & 2.8151 & 4.0351 & 2.1400\\
    \midrule
    Dropout (0.3) & 0.9230 & 0.7203 & 0.8811 & 0.6921 & 1.2802 & 0.9210 & 4.2674 & 2.8179 & 3.9786 & 2.1514 \\
    NoDrop & 0.9236 & 0.7218 & 0.8821 & 0.6928 & 1.2729 & 0.9051 & 4.2831 & 2.8177 & 3.9879 & 2.1172\\
    \midrule
    Vanilla CL & 0.9306 & 0.7253 & 0.8794 & 0.6890 & 1.3326 & 0.9437 & 4.3020 & 2.8621 & 3.9631 & 2.1689\\
    No GLCL & 0.9257 & 0.7217 & 0.8806 & 0.6912 & 1.2812 & 0.9046 & 4.2950 & 2.8314 & 3.9967 & 2.1429\\
    \bottomrule
    \end{tabular}
\end{table}
\label{sec:ablation_study}
This study comprehensively investigates the impact of all components of the model and their variants on the overall performance of DCGC.

\paragraph{Effects of Dual-Attention.} In DCGC, we propose a dual-attention mechanism to adaptively identify both informative convolutional output channels and discriminative ``aligned'' features. As shown in Table \ref{tab:ablation}, overall, when introducing single attention (channel/feature), the RMSE drops by $0.23\% \sim 2.49\%$ and the MAE drops by $0.77\% \sim 2.71\%$. These results indicate that explicitly reweighting informative experts or feature dimensions helps the model suppress less useful responses and focus on more predictive interaction patterns. When replacing single attention (channel/feature) with our dual-attention mechanism, the RMSE further drops by $0.08\% \sim 2.19\%$ and the MAE further drops by $0.11\% \sim 2.45\%$.  Notably, a single attention branch sometimes dominates the performance gain on some datasets (e.g., feature attention on GZspeed), which can also be observed in Sec \ref{sec:para_ana}.

\paragraph{Effects of Personalized Gating.} 
In the Multi-gate Mixture-of-Experts framework, existing methods typically introduce dropout operations to mitigate the inherent expert polarization phenomenon \cite{zhao201mmoe_dropout}. Therefore, we compare the impact of dropout and our personalized gating mechanism here. As shown in Table \ref{tab:ablation}, replacing dropout with our personalized gating mechanism, the RMSE drops by $0.48\% \sim 1.82\%$ and the MAE drops by $0.52\% \sim 2.27\%$. What's more, due to the severe data sparsity in Beauty, directly introducing dropout severely degrades the overall performance, while our method achieves consistent  improvements.

\begin{figure}[h]
    \centering
    \begin{subfigure}{0.48\linewidth}
        \centering
        \includegraphics[width=\linewidth]{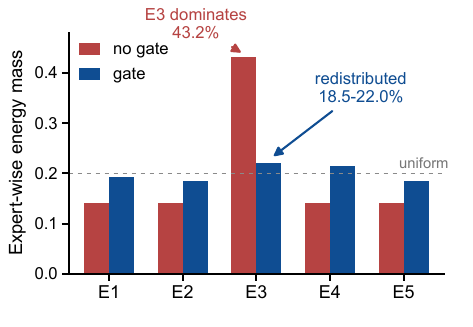}
        \caption{Mov100k}
        \label{fig:example_a}
    \end{subfigure}
    \hfill
    \begin{subfigure}{0.48\linewidth}
        \centering
        \includegraphics[width=\linewidth]{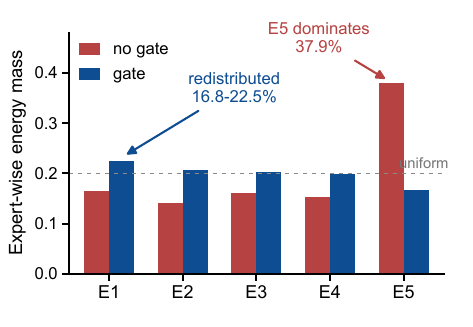}
        \caption{GZspeed}
        \label{fig:example_b}
    \end{subfigure}
    \caption{Effect of personalized gating on expert load distribution. The bars report the expert-wise energy mass defined in Eq.~\eqref{eq:expert_energy_mass} for the non-gated and gated variants. The dashed line indicates the uniform load level with five experts. A peaked distribution indicates stronger expert-response polarization, whereas a flatter distribution suggests more balanced expert utilization.}
    \label{fig:expert_distribution}
\end{figure}
To examine whether the personalized gating mechanism improves expert load
balancing, we define the expert-wise energy mass using a generic response
representation \(\Z\in\R^{C\times R}\). For each tensor entry, the
response energy assigned to the \(c\)-th convolution expert is computed by
aggregating the absolute response values over the feature-rank dimension:
\begin{equation}
    \eta_c(\Z) =
    \sum_{r=1}^{R}
    \left|Z_{c,r}\right|,
    \quad c=1,2,\dotsm,C.
    \label{eq:expert_energy}
\end{equation}
where $C$ is the number of convolution experts and $R$ is the feature-rank
dimension. The normalized expert-wise energy mass is then defined as
\begin{equation}
    \rho_c(\Z)
    =
    \frac{\eta_c(\Z)}{\sum_{i=1}^{C}\eta_i(\Z)}
    =
    \frac{
    \sum_{r=1}^{R}
    \left|Z_{c,r}\right|
    }{
    \sum_{i=1}^{C}
    \sum_{r=1}^{R}
    \left|Z_{i,r}\right|
    }.
    \label{eq:expert_energy_mass}
\end{equation}
Here, \(\rho_c(\Z)\) measures the proportion of response energy assigned to the
\(c\)-th convolution expert. In the visualization, \(\rho_c(\Z)\) is first
computed for each evaluated tensor entry and then averaged over all evaluated
entries.

In the non-gated setting, the personalized gating operation in
Eq.~\eqref{eq:personalized_gating} is removed, so the expert load is computed by
setting \(\Z=\E^{\rm ng}\) in Eq.~\eqref{eq:expert_energy_mass}, where
\(\E^{\rm ng}\) follows the dual-attention response in
Eq.~\eqref{eq:dual_attention_response}. In the gated setting, we set
\(\Z=\X_{\rm att}^{\rm g}\), where \(\X_{\rm att}^{\rm g}\) is obtained from
Eq.~\eqref{eq:personalized_gating}. The resulting quantities,
\(\rho_c(\E^{\rm ng})\) and \(\rho_c(\X_{\rm att}^{\rm g})\), compare expert
loads before and after personalized gating. As shown in
Fig.~\ref{fig:expert_distribution}, the non-gated model concentrates response
energy on a dominant expert: E3 accounts for \(43.2\%\) of the total expert
energy on Mov100k, and the dominant expert on GZspeed accounts for \(37.9\%\).
In contrast, the gated model redistributes the energy more evenly across
experts, with masses ranging from \(18.5\%\) to \(22.0\%\) on Mov100k and from
\(16.8\%\) to \(22.5\%\) on GZspeed. These results indicate that personalized
gating mitigates expert-response polarization and encourages a more balanced
use of the expert subspace.

\paragraph{Effects of Group-Level Contrastive Learning.}  We investigate the impact of two variants on model performance: 1) ablation of the proposed contrastive learning module; 2) replacement with vanilla contrastive learning strategy, which simply takes observed samples as positive samples and unobserved items as negative samples. As shown in Table \ref{tab:ablation}, due to the substantial noise in unobserved data, adopting BPR as an auxiliary task degrades model performance on four evaluated datasets. In contrast, introducing our proposed GLCL reduces RMSE by $0.42\% \sim 2.11\%$ and MAE by $0.21\% \sim 2.02\%$ across all datasets.
\begin{figure}[t]
    \centering
    \begin{subfigure}[b]{0.48\textwidth}
        \centering
        \includegraphics[width=0.48\linewidth]{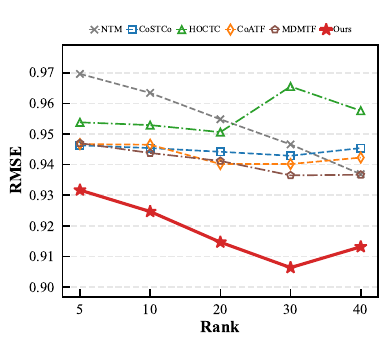}
        \hfill
        \includegraphics[width=0.48\linewidth]{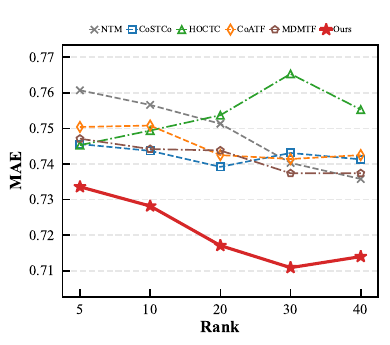}
        \caption{Influence of Rank on Mov100k}
        \label{fig:rank_mov}
    \end{subfigure}
    \hfill
    \begin{subfigure}[b]{0.48\textwidth}
        \centering
        \includegraphics[width=0.48\linewidth]{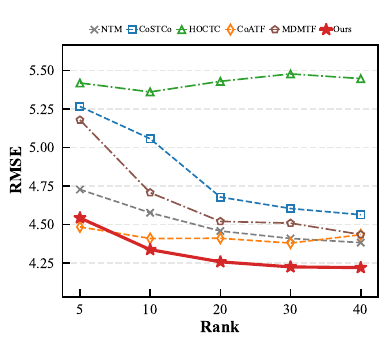}
        \hfill
        \includegraphics[width=0.48\linewidth]{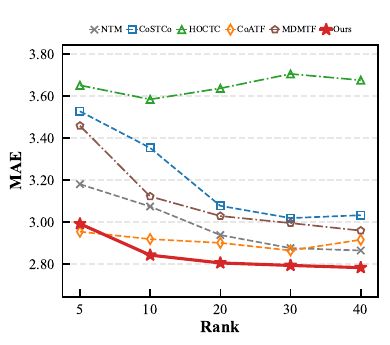}
        \caption{Influence of Rank on GZspeed}
        \label{fig:rank_gz}
    \end{subfigure}
    
    \vspace{0.5cm} 
    
    \begin{subfigure}[b]{0.48\textwidth}
        \centering
        \includegraphics[width=0.48\linewidth]{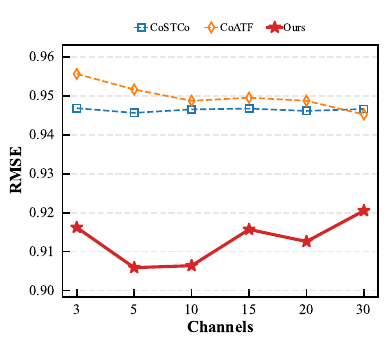}
        \hfill
        \includegraphics[width=0.48\linewidth]{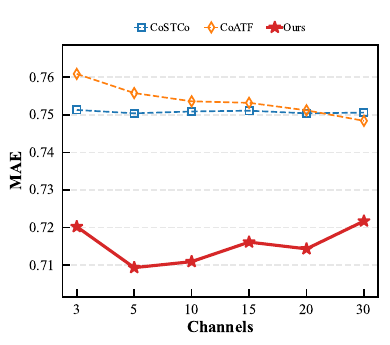}
        \caption{Influence of Channels on Mov100k}
        \label{fig:chan_mov}
    \end{subfigure}
    \hfill
    \begin{subfigure}[b]{0.48\textwidth}
        \centering
        \includegraphics[width=0.48\linewidth]{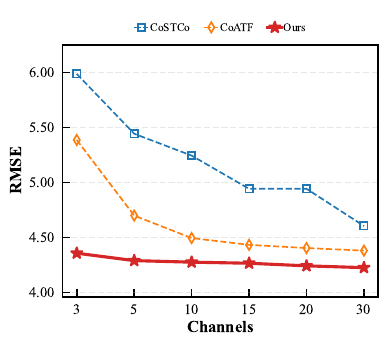}
        \hfill
        \includegraphics[width=0.48\linewidth]{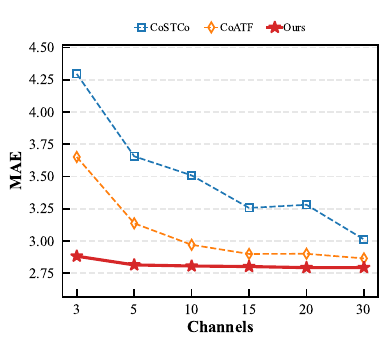}
        \caption{Influence of Channels on GZspeed}
        \label{fig:chan_gz}
    \end{subfigure}
    \caption{Parameter sensitivity analysis on Mov100k and GZspeed datasets. 
    The first row (a)-(b) illustrates the impact of \textbf{Rank}, while the second row (c)-(d) shows the impact of \textbf{Channels}.}
    \label{fig:sensitivity_analysis}
\end{figure}

\subsection{Parameter Analysis}
\label{sec:para_ana}
Figure \ref{fig:rank_mov} and Figure \ref{fig:rank_gz} present the error of different methods on both spatiotemporal and recommendation datasets by varying the factorization rank from 5 to 40. Due to the introduction of feature attention and group-level contrastive learning, our method is quite competitive for all target ranks and shows a stable performance improvement with increasing ranks.

In Figures \ref{fig:chan_mov} and \ref{fig:chan_gz}, we analyze the impact of the number of output channels in the convolutional layers on the performance of three CNN-based methods across both types of tensors. We vary the channel count from 3 to 30, with 30 corresponding to the configuration employed in CoSTCo and CoATF. The results demonstrate that our model maintains highly competitive performance even with a limited number of convolutional channels, particularly in relatively complex recommendation scenarios. Notably, since the computational complexity of convolution layers grows rapidly with the number of channels, the incorporation of channel attention enables our model to significantly reduce the computational burden associated with convolution operations while simultaneously enhancing performance.

\subsection{Discussion of Limitations}
Although our model achieves competitive performance, two limitations remain for
future investigation. First, the current dual-attention module fuses expert-level and feature-level attention scores using a simple additive operation. This strategy is effective and lightweight, but may not fully capture the complementary information between expert-level and feature-level importance. More adaptive fusion mechanisms could further improve the information utility of the dual-attention module. Second, GLCL currently relies on predefined feedback intervals for contrastive supervision. Since the optimal interval partition may vary across datasets and sparsity levels, future work could explore adaptive interval construction based on feedback distribution or interaction density.
 
\section{Conclusion}
This paper proposes DCGC, a neural tensor factorization method for sparse tensor completion that models cross-mode latent interactions with a fine-grained mixture of convolutional experts and alleviates sparsity through group-level contrastive learning. Experiments on five traffic and recommendation datasets show that DCGC outperforms tensor factorization baselines and remains competitive with specialized GNN-based methods for sparse recommendation. 
In future work, we are interested in extending our method to higher-dimensional data scenarios.

\begin{credits}
\subsubsection{\ackname} 
This study was supported by the National Key Research and Development Program of China (No.~2025YFA1018600), the National Natural Science Foundation of China (via fund NSFC-12571561), and the Fundamental Research Support Program of HUST (2025BRSXB0004).
\end{credits}
%
%
%
\bibliographystyle{splncs04}
\bibliography{mybibliography}
\end{document}